\documentclass[10pt,twocolumn,letterpaper]{article}

\usepackage[pagenumbers]{wacv} % To force page numbers, e.g. for an arXiv version
\usepackage{booktabs}
\usepackage{array, multirow}
\usepackage{tablefootnote}
\usepackage[table]{xcolor}
\usepackage{caption}
\usepackage{subcaption}
\usepackage{wrapfig}
\usepackage{float}
\usepackage{times}
\usepackage{epsfig}
\usepackage{graphicx}
\usepackage{amsmath}
\usepackage{amssymb}
\makeatletter
\robustify\@latex@warning@no@line
\makeatother
\usepackage{authblk}

\usepackage[accsupp]{axessibility}  % Improves PDF readability for those with disabilities.

\usepackage[pagebackref,breaklinks,colorlinks]{hyperref}
\newcommand*{\threeemdash}{\rule[0.5ex]{2em}{0.55pt}}
\newcommand{\ourmodel}{ProS}

\def\wacvPaperID{124} % *** Enter the WACV Paper ID here
\def\confName{WACV}
\def\confYear{2024}

\begin{document}
	%%%%%%%%% TITLE
	
	\title{\ourmodel{}: Facial Omni-Representation Learning via Prototype-based Self-Distillation\vspace{-1em}}
	
	\author[1]{Xing Di}
	\author[1]{Yiyu Zheng}
	\author[2]{Xiaoming Liu}
	\author[3]{Yu Cheng}
	\affil[1]{ProtagoLabs Inc.}
	\affil[2]{Michigan State University} 
	\affil[3]{Rice University}
	\affil[ ]{{\tt\small \{xing.di,yiyu.zheng\}@protagolabs.com, liuxm@cse.msu.edu, yu.cheng@rice.edu}}
	
	% \author{Xing Di, Yiyu Zheng\\
		% ProtagoLabs Inc.\\
		% % 8221 Courthouse RD, Vienna, VA,22182 \\
		% {\tt\small {\{xing.di,yiyu.zheng\}@protagolabs.com}}
	% \and
	% Xiaoming Liu\\
	% Michigan State University\\
	% % 8221 Courthouse RD, Vienna, VA,22182 \\
	% {\tt\small liuxm@cse.msu.edu}
	% \and
	% Yu Cheng\\
	% The Chinese University of Hong Kong\\
	% % First line of institution2 address\\
	% {\tt\small chengyu@cse.cuhk.edu.hk}
	% }

\twocolumn[{%
	\renewcommand\twocolumn[1][]{#1}%
	\maketitle
	\begin{center}
		\vspace{-3em}
		\centering
		\captionsetup{type=figure}
		\includegraphics[width=.98\textwidth,height=5cm]{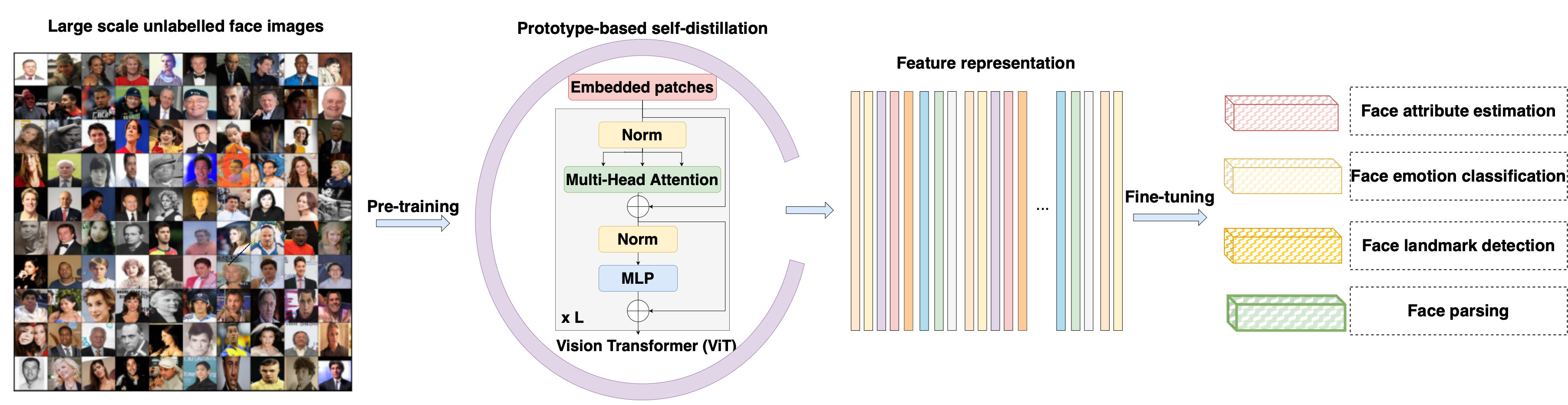}
		\captionof{figure}{ This paper presents a pre-training model that learns facial omni-representations via a \textbf{pro}totype-based \textbf{s}elf-distillation (\ourmodel{}) . For pre-training, \ourmodel{} learns a general face representation from given large-scale \textbf{unlabeled} face images. Afterwards, the learned omni-representaion  can be conveniently utilized in multiple downstream tasks by simple fine-tuning.
		}
		\label{fig:framework}
	\end{center}%
}
]

\maketitle
\thispagestyle{empty}

%%%%%%%%% ABSTRACT
\begin{abstract}
	% Current supervised face representation learning requires a large amount of training data. Collecting such large-scale real-world face images is challenging due to annotation efforts and privacy issues. 
	% To address these issues, we propose the \textbf{pro}totype-based \textbf{s}elf-distillation (\ourmodel{}), which aims to learn facial omni-representation from a large amount of \textbf{unlabeled} face images. 
	% In particular, \ourmodel{} consists of two vision-transformers (teacher and student models) that are trained with different augmented images (cropping, blurring, coloring, etc.). 
	% Besides, we build a face-aware retrieval system along with augmentations to obtain the curated images by eliminating most non-face images. We observe there are many variations in the augmented images. To boost the discrimination of learned features, we propose another prototype-based matching loss by matching the similarity distributions between features (teacher or student) and a set of learnable prototypes. 
	% After pre-training, the teacher vision transformer from \ourmodel{} is utilized as the backbone to a couple of downstream tasks such as attribute estimation, expression recognition, and landmark alignment by simply fine-tuning with the additional layers. 
	% Extensive experiments demonstrate our method can achieve state-of-the-art performances on most tasks, in both full and few-shot settings. 
	% Additionally, we also conduct the pre-training with synthetic face images, where \ourmodel{} still achieves promising performances.
	This paper presents a novel approach, called Prototype-based Self-Distillation (\ourmodel{}), for unsupervised face representation learning.
	The existing supervised methods heavily rely on a large amount of annotated training facial data, which poses challenges in terms of data collection and privacy concerns.
	To address these issues, we propose \ourmodel{}, which leverages a vast collection of unlabeled face images to learn a comprehensive facial omni-representation.
	In particular, \ourmodel{} consists of two vision-transformers (teacher and student models) that are trained with different augmented images (cropping, blurring, coloring, etc.).
	Besides, we build a face-aware retrieval system along with augmentations to obtain the curated images comprising predominantly facial areas. 
	To enhance the discrimination of learned features, we introduce a prototype-based matching loss that aligns the similarity distributions between features (teacher or student) and a set of learnable prototypes.
	After pre-training, the teacher vision transformer serves as a backbone for downstream tasks, including attribute estimation, expression recognition, and landmark alignment, achieved through simple fine-tuning with additional layers.
	Extensive experiments demonstrate that our method achieves state-of-the-art performance on various tasks, both in full and few-shot settings.
	Furthermore, we investigate pre-training with synthetic face images, and \ourmodel{} exhibits promising performance in this scenario as well.
\end{abstract}

%%%%%%%%% BODY TEXT
\section{Introduction}\label{sec:intro}

\begin{figure*}[t]
	\centering
	\includegraphics[width=0.98\linewidth]{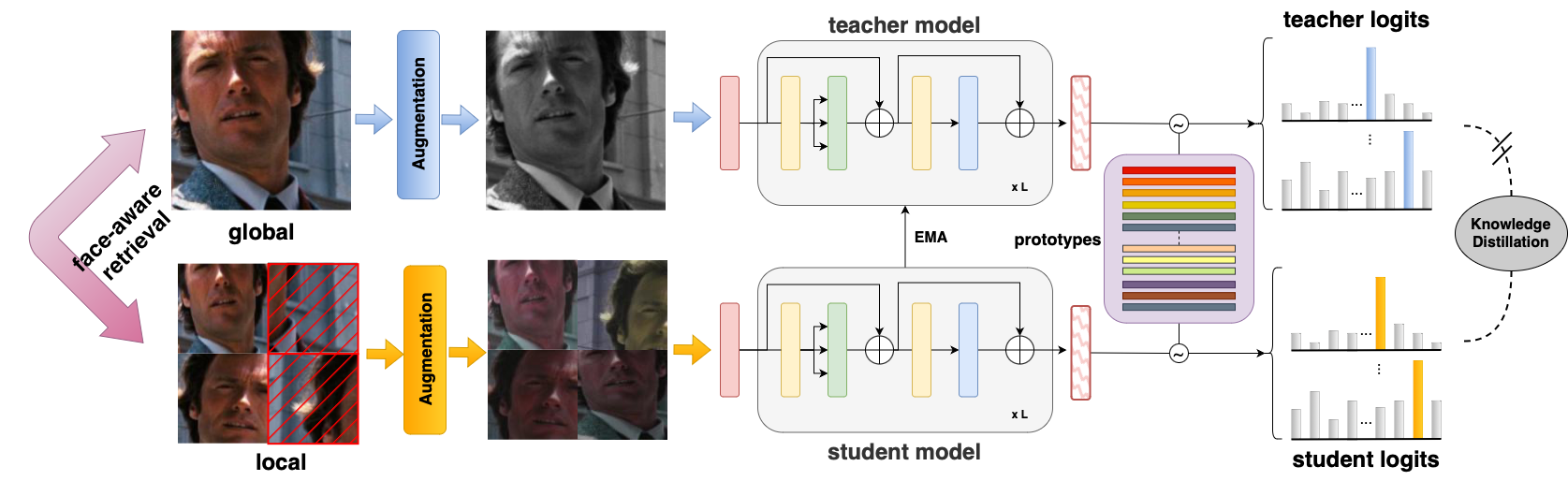}
	\caption{The proposed prototype-based self-distillation method. There are two branches: global and local. The global and local images are obtained through the multi-crops \cite{caron2020unsupervised,caron2021emerging}. To obtain the curated face images in local branch, we propose a face-aware retrieval system followed by the augmentations. The teacher and student models have the same vision-transformer architectures (ViT-S/16) but with different parameters. The self-knowledge-distillation between the student and teacher features is penalized via the similarity distribution between features and the learnable prototypes. By this, the networks are forced to leverage the mutual semantics between local and global views.}
	\label{fig:method}     
\end{figure*}

%  we discuss face representation learning 
Learning good face representation is crucial for face analysis tasks such as face recognition~\cite{ranjan2017l2,liu2017sphereface,liu2016large,wang2017normface,wang2018additive,wang2018cosface,deng2019arcface,schroff2015facenet,nech2017level}, attribute estimation \cite{cao2018partially,liu2015deep,shu2021learning}, expression classification~\cite{wen2021distract,zhang2022learn}, landmark localization \cite{huang2021adnet,kumar2020luvli,wang2019adaptive}. Among these tasks, existing state-of-the-art (SoTA) methods own their success not only to the sophisticated network design but also large-scale training datasets. However, acquiring manually-annotated large-scale facial images is expensive and difficult for large labor-work and privacy issues \cite{Guo2016MSCeleb1MAD}. For instance, it is hard to obtain the consent of all involved identities for face recognition datasets.

% we discuss the pre-training here to avoid the large labor annotation, 
Recently, self-supervised learning has gained intensive interest due to the remarkable success of training generalizable models in both natural language processing \cite{devlin2018bert,radford2018improving,radford2019language,brown2020language} and computer vision \cite{he2022masked,chen2020simple,he2020momentum,grill2020bootstrap,caron2021emerging,assran2022masked,xie2022simmim,wei2022masked,oquab2023dinov2}. Such a pre-trained model has shown the following advantages: (a) the learned feature shows superiority on transfer-learning especially in few-shot settings, where it achieves a promising performance when data acquisition is limited. (b) the learned model is scalable for further development on diverse downstream tasks.  To our acknowledgment, only few works \cite{zheng2022general,bulat2022pre} explored the semi/self-supervised learning on face model. FaRL \cite{zheng2022general} explored a contrastive loss and masked image modeling for learning features from image-text pairs. FRL \cite{bulat2022pre} learned the face representation based on ResNet \cite{he2016deep} via SwAV \cite{caron2020unsupervised}.

% we discuss \ourmodel{} here
Different from those previous methods,  we propose a vision-transformer framework for learning face representation via \textbf{pro}totype-based \textbf{s}elf-distillation (\ourmodel{}). As shown in Figure~\ref{fig:framework}, \ourmodel{} aims to learn the advanced feature representations given large-scale face images \textbf{without} labeling. In particular, \ourmodel{} is trained in self-knowledge-distilling via a local-to-global manner. Our work is inspired by DINO \cite{caron2021emerging} but with  (i) a modified sample-to-prototype matching loss and (ii) a proposed face-aware retrieval system for curated data augmentation. 

As shown in Figure~\ref{fig:method}, the global and local images are obtained from the same input image via multi-crop \cite{caron2021emerging,caron2020unsupervised} followed by the proposed face-aware retrieval system. The face-aware retrieval system aims to filter out most non-face images. Afterward, the curated images are fed into two sets of separate augmentations. The augmented images in global and local views are given to the teacher and student models respectively. By matching the features extracted from the teacher and student models, the loss gradient is back-propagated to the student model only for updating parameters. The parameters of the teacher model are updated through the exponential moving average \cite{grill2020bootstrap}.

\begin{figure}[t]
	\centering
	\includegraphics[width=0.9\linewidth]{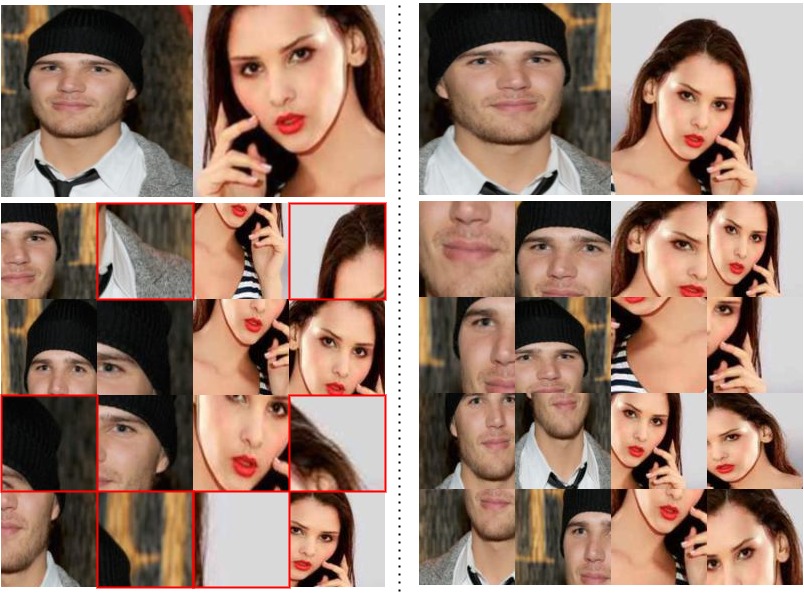} \\
	\hskip 0 pt (a) uncurated  \hskip 40pt (b) curated 
	\vskip -5 pt 
	\caption{The multi-cropped samples from MS1M \cite{Guo2016MSCeleb1MAD}. We compare the (a) uncurated and (b) curated local samples with/without the facial-retrieval system. The global images are shown on top for reference. The non-face images (with red bounding boxes) are deduplicated.}
	\label{fig:low-quality}
\end{figure}

% we discuss the data curation here
During training, we observed that there are certain amounts of non-face images obtained in the local view as highlighted by the red bounding boxes in Figure~\ref{fig:low-quality}(a). Recent studies show those unidentifiable images are detrimental to the training procedure \cite{kim2022adaface}. To eliminate those outliers, we build a face-aware retrieval system. In particular, we compute the face embeddings using the pre-trained ArcFace \cite{deng2019arcface} for both the global and local images. The cosine similarity is used as a distance measure between local and global images. We displace those local face images with similarities lower than a threshold. We demonstrate two curated samples for the same input images in Figure~\ref{fig:low-quality}(b).

% we discuss the prototypes here
After the face-aware retrieval system, we find there are lots of variations in the curated images.  To boost the feature discrimination, we introduce a set of learnable prototypes during training. Instead of directly computing the similarity of the features between the teacher and student samples, we compute two sample-to-prototype distributions: one between teacher samples to the prototypes, and the other one between student samples to the prototypes. We use the differences between these two distributions to penalize the model training. In this manner, both the prototypes and teacher-student models are optimized in every iteration where the features try to get close to positive prototypes and keep away from negative prototypes \cite{deng2021variational}. In addition,  to mitigate the privacy issue of using real face images, we also explore synthesized face images for learning face representations.  In particular, we simply train a StyleGAN2 \cite{karras2020training,Karras2019stylegan2} from scratch on MS1M \cite{Guo2016MSCeleb1MAD}. The synthetic data is generated via  randomly-sampled noise from a normal distribution.  

% % we discuss the synthetic data here
% In addition,  to mitigate the privacy issue of using real face images, we also explore synthesized face images for learning face representations.  In particular, we simply train a StyleGAN2 \cite{karras2020training,Karras2019stylegan2} from scratch on MS1M \cite{Guo2016MSCeleb1MAD}. The synthetic data is generated via  randomly-sampled noise from a normal distribution.  
% We will release all the pre-training and fine-tuning codes as well as the synthetic data/generator upon acceptance. We hope these could serve an important role, such as the dataset and methods, for further research.

To this end, our contributions in this paper can be summarized as follows:
\begin{itemize}
	\item We propose a novel pre-training framework (\ourmodel{}) for learning facial omni-representation from large-scale face images without labeling. A learnable prototype-based matching loss and a face-aware retrieval system are introduced along with \ourmodel{}. 
	\item We conduct extensive experiments for evaluating \ourmodel{} on various face analysis tasks. Our proposed \ourmodel{} can achieve state-of-the-art results over different baselines on all the tasks in few-shot settings.
	\item We explore the capability of \ourmodel{} on synthetic face images. To our best knowledge, \ourmodel{} is the first to work on self-supervised pre-training on large-scale synthetic face images. We show that our method still obtains promising performances.
\end{itemize}

\section{Related work}
We walk through the related literature on self-supervised training, facial representation learning, and face synthesis in this section. 

\subsection{Self-supervised training}
Self-supervised learning methods \cite{chen2020simple,he2022masked,xie2022simmim,caron2020unsupervised,caron2021emerging,bardes2022vicreg,zhou2022image,oquab2023dinov2,chen2020uniter} have gained remarkable attention as an effective unsupervised learning strategy  for learning robust image representations and eliminating the need to annotate vast quantities of data manually. For instance,  SimCLR \cite{chen2020simple} maps the initial embeddings from two augmented views of an image into another space where the infoNCE loss is applied to encourage similarity between the views.
DINO \cite{caron2021emerging} feeds two different views of an image into the teacher and student encoders and maps the student network's weights to the teacher's by a moving average. SWAV \cite{caron2020unsupervised} simultaneously clusters the data while enforcing consistency between cluster assignments when given different augmented views of an image. In addition, MAE \cite{he2022masked} and SimMIM \cite{xie2022simmim} are two concurrent masked image modeling (MIM)  that directly reconstruct masked image patches.

\subsection{Facial representation learning}
Existing face representation learning methods can be categorized into two classes: the proxy-based learning \cite{ranjan2017l2,liu2017sphereface,liu2016large,wang2017normface,wang2018additive,wang2018cosface,deng2019arcface} and pair-wise learning \cite{schroff2015facenet,nech2017level,Guo2016MSCeleb1MAD}. As the class labels are known, proxy-based learning aims to optimize the similarity between given samples and a set of proxies representing each class. On the one hand, those methods \cite{deng2019arcface,wang2018cosface,wang2018additive,liu2017sphereface,chen2017noisy,sun2020circle,huang2020curricularface,kim2022adaface} put a margin penalty into the softmax loss and a global comparison between samples and proxies is conducted. On the other hand, those methods \cite{oh2016deep,rippel2016metric,schroff2015facenet,wu2017sampling,sohn2016improved,shi2020towards,Wang2016WalkAL} involve different triplet strategies (representation selection, hard-mining) in mini or larger batch to leverage the underline pairwise information. 

Recently, a few studies have been done for face analysis on few-shot \cite{browatzki20203fabrec}, weakly-supervised  \cite{zheng2022general}, and self-supervised learning \cite{bulat2022pre,wiles2018self,cai2022marlin}.  For instance, Browatzki \etal  \cite{browatzki20203fabrec} proposed a few-shot face alignment framework with an image reconstruction by an auto encoder-decoder. Zheng \etal \cite{zheng2022general} proposed the FaRL for pre-training  the vision-transformer model by leveraging the semantics between web-text and face image pairs.  Wiles \etal \cite{wiles2018self} introduced a self-supervised manner for predicting the motion field between two facial images to learn efficient face representations. Vielzeuf \etal \cite{vielzeuf2019towards} introduced a common embedding for multi-source features from different trained models by an auto-encoding framework. Bulat \etal
introduced the unsupervised training model FRL \cite{bulat2022pre} for pre-training ResNet on the collected large-scale dataset. 
% Cai \etal \cite{cai2022marlin} designed a masked autoencoder model to learn facial embeddings from non-annotated facial videos.

% Particularly, the most pertinent study (the FRL model \cite{bulat2022pre}) to our work utilized  SwAV \cite{caron2020unsupervised} for unsupervised pre-training ResNet \cite{he2016deep} model. The main differences between Two main differences between our work and FRL are (i) the vision-transformer architecture and (ii) the momentum encoder. 
% \subsection{Self-supervised training}
% Self-supervised learning methods \cite{he2022masked,assran2022masked,xie2022simmim,wei2022masked,caron2021emerging,chen2021exploring} have gained remarkable attention as an effective unsupervised learning strategy  for learning robust image representations and eliminating the need to annotate vast quantities of data manually. 
% While only a few works \cite{bulat2022pre,wiles2018self} focused on face images and showed the features learned can be effective for downstream tasks with few labels.  For instance, Wiles \etal \cite{wiles2018self} introduced a self-supervised manner for predicting the motion field between two facial images to learn efficient face representations. Bullet \etal \cite{bulat2022pre} explored the self-distillation strategy SwAV \cite{caron2020unsupervised} to boost face recognition and several face analysis tasks with limited data.  Vielzeuf \etal \cite{vielzeuf2019towards} introduced a common embedding for multi-source features from different trained models by an auto-encoding framework.

\subsection{Face synthesis}
With the remarkable ability of GANs \cite{goodfellow2014generative,goodfellow2020generative}, face synthesis has seen rapid developments, such as StyleGAN \cite{karras2019style} and its variations \cite{Karras2019stylegan2,karras2020training,karras2021alias} which can generate high-fidelity face images from random noises. Synthetic face data has shown significant improvements on various tasks such as learning pose-invariant models \cite{tran2017disentangled,zhao20183d,yin2017towards,di2021heterogeneous},  cross-spectrum models \cite{di2019polarimetric,di2018polarimetric,yu2021lamp,poster2021large} as well as reducing data bias \cite{kortylewski2019analyzing,qiu2021synface,ruiz2020morphgan,shi2020towards}. Unlike previous methods, we explore the possibility of using synthesized face images for self-supervised pre-training. We hope the synthetic data could be used as an alternative to real face images to  avoid privacy issues when collecting data.

\section{Proposed method}

The proposed prototype-based self-distillation pre-training method is illustrated in Figure~\ref{fig:method}. In particular, it contains two branches: the global and the local. The global and local images are multi-cropped from the same original image, resized at different scales and fed into the teacher and student models separately to obtain the corresponding features. 

During the experiments, we find there are some non-face images cropped from the local branch which could diminish the discrimination of the learned features. Therefore, a face-aware retrieval system has been built for eliminating non-face images. Specifically, we utilize the pre-trained Arcface \cite{deng2019arcface} to extract the features from both the global and local images. We filter out the local images that have a lower cosine distance between local-global features. In this paper, we set the distance threshold $\theta = -0.5 \in [-1.0, 1.0]$ by visual inspection of the retrieval results.

Inspired by recent findings \cite{shi2019probabilistic,kim2022adaface} that a "high" similarity score would be obtained from features of low-quality face images, a set of prototypes is utilized for penalizing the knowledge distillation. During each training iteration, the similarity scores are calculated between the teacher/student image features and these memorized prototype features instead of directly between the teacher/student features themselves. The cross-entropy loss is calculated between the similarity vectors. The student network parameters are optimized by back-propagating the loss gradient while the teacher network parameters are updated via an exponential moving average (EMA) of the student parameters \cite{grill2020bootstrap}. Additionally, the prototypes are also optimized along with the network parameters by backpropagation.

\subsection{Prototype-based self-distillation}
Given a large-scale collection of unlabeled face images, the knowledge distillation aims to train the teacher and student models,  parameterized as $\theta_{t}, \theta_{s}$,  for matching the output features $f_{\theta_{t}}(x)$ and $f_{\theta_{s}}(x)$  given an input image $\mathbf{x}$. 

In each iteration, we sample a mini-batch of $B$ images $\{\mathbf{x}_{i}\}_{i=1}^{B}$. The global images $\{\mathbf{x}_{m \rightarrow i}^{g}\}_{m=1}^{M}$ are obtained by a random crop of $i$-th original images followed by a set of global augmentations. Similarly, the local images $\{\mathbf{x}_{n \rightarrow i}^{l}\}_{n=1}^{N}$ are based on another random crop of $i$-th original image followed by a set of local augmentations. For brevity, we omit the captions as $\mathbf{x}_{m}^{g}$ and $\mathbf{x}_{n}^{l}$.   Following prior work \cite{dosovitskiy2021an}, we "patchify" the input images into a set of sequential patches without overlapping. After, the global and corresponding local patches are fed into the teacher and student models respectively.

Let $f_{\theta_{t}}(\mathbf{x}_{m}^{g}) \in  R^{d}$ and $f_{\theta_{s}}(\mathbf{x}_{n}^{l}) \in  R^{d}$ denote the $d$-dim feature vectors obtained from teacher network and student network respectively. Additionally, a set of the learnable prototypes is denoted as $\mathbf{p}\in R^{K\times d}$. Instead of directly utilizing the teacher and student features, we use the cosine similarity between the student/teacher feature and these prototypes as the features.
The prediction is calculated as follows:
\begin{equation}\label{similarity}
	\resizebox{0.9\hsize}{!}{	
		$s_{n}^{l} = softmax( \frac{\mathbf{p} \cdot f_{\theta_{s}}(\mathbf{x}_{n}^{l})}{\tau_{l}} );  s_{m}^{g} = softmax( \frac{\mathbf{p} \cdot f_{\theta_{t}}(\mathbf{x}_{m}^{g})}{\tau_{g}} ),$
	} 
\end{equation} 

where $\cdot$ denotes the dot product and $\tau_{g} \in (0,1), \tau_{l} \in (0,1)$. All the output features are $L_{2}$ normalized to mitigate the scale influence. To prevent the model collapse, we choose $\tau_{g}$ to be smaller than $\tau_{l}$, and the global sharpening is utilized by $softmax$ during training. The training objective is defined as follows:
\begin{equation}\label{eq:training objective}
	\mathcal{L} = \frac{1}{B}\sum_{i=1}^{B} \mathcal{H}_{i}(s_{m}^{g}, s_{n}^{l}) - \mathcal{H}(\tilde{s}^{l}),
	% 	\mathcal{L} = \frac{1}{M\cdot N}\sum_{m=1}^{M}\sum_{n=1}^{N} \mathcal{H}(s_{m}^{g}, s_{n}^{l}) - \mathcal{H}(\tilde{s}^{l}), 
\end{equation}
where $\mathcal{H}_{i}(s_{m}^{g}, s_{n}^{l})$ is the $i$-th cross-entropy between $s_{m}^{g}, s_{n}^{l}$  as Eq~\eqref{eq:cross entropy}:
\begin{equation}\label{eq:cross entropy}
	\mathcal{H}_{i}(s_{m}^{g}, s_{n}^{l}) =  
	\frac{1}{M\cdot N}\sum_{m=1}^{M}\sum_{n=1}^{N} \mathcal{H}(s_{m}^{g}, s_{n}^{l}),
\end{equation}
while, $\mathcal{H}(\tilde{s}^{l})$ is the entropy regularization \cite{bardes2021vicreg} as Eq~\eqref{eq:entropy}
\begin{equation}\label{eq:entropy}
	\mathcal{H}(\tilde{s}^{l}) =   \frac{1}{B\cdot D} \sum_{i=1}^{B}\sum_{n=1}^{N}s_{n\rightarrow i}^{l},
\end{equation}

\subsection{Model architecture}

The teacher and student models are the vision-transformer encoders \cite{dosovitskiy2021an,touvron2021training}. For a fair comparison with other face analysis tasks, ViT-S/16 \cite{touvron2021training} is chosen, whose number of parameters is similar to the common ResNet-50 \cite{he2016deep} (21M {\it vs} 23M). Specifically, ViT-S/16 is a $12$-layer $384$-width visual transformer with $224 \times 224$ resolution input. In our work, the global input images are $224 \times 224$ while the local images are $96\times 96$. These global/local images are firstly split into $14\times 14$ and $6\times 6$ patches respectively. Thus, one learnable $cls$ token is prepended to the $196$/$36$ embedding. In pre-training, additional $3$ fully-connected layers are added as the projector to the output transformer for further optimization by Eq.~\eqref{eq:training objective}. The prototypes are a set of learnable variables with random initialization. We set the output feature dimension as 256  as  the prototypes.

\subsection{Pre-training details} 
The teacher and student models are trained from scratch with randomly initialized parameters. The total training runs $20$ epochs with a total batch size $64 \times 4$ on $4$ Nvidia 3090 GPUs. The AdamW optimizer is utilized with weight decay as $0.04$. The initial learning rate is $0.0002$ with $2$ warmed-up epochs to $0.001$ and then cosine decay to $1\text{e}-6$ in the next $18$ epochs. The teacher/student temperatures are set as $0.025$ and $0.1$. The number of prototypes is set to  $1,024$. The learnable prototypes are randomly initialized with a uniform distribution between $[-1/\sqrt{d},1/\sqrt{d}]$, where $d$ is the output dimension here. The prototypes are updated iteratively along with the network parameters by back-propagation. To make use of all the unlabeled images, we directly input the raw face images for pre-training without any further preprocessing like face detection, cropping, or alignment.

% In practice, we find that using the detected face images (the aligned version of MS1M dataset) leads to an unstable training loss. We want to leave the exploration of combining face detection and multi-crop strategy in future work.

\subsection{Synthetic data}
\begin{figure}
	\centering
	\includegraphics[width=0.98\linewidth]{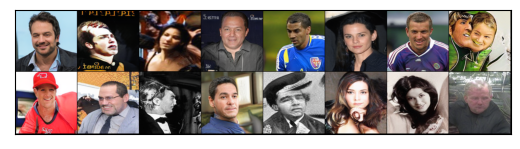}
	\vskip -10 pt
	\caption{Samples of synthetic face images.}
	\label{fig:synsamples}
	\vskip -10 pt
\end{figure}

We train the original StyleGAN2 \cite{Karras2019stylegan2} on MS1M dataset \cite{Guo2016MSCeleb1MAD} to obtain the synthetic data. In particular, the images are all resized to $256 \times 256$ to train the adversarial generative networks. The "paper256" setting \cite{karras2020training} is utilized. When training on different data sizes, images are randomly selected. Once the training is completed, the synthetic face images are obtained through the corresponding generator via input noise vectors, which are sampled from a standard normal distribution.  Samples of the synthesized face images are shown in Figure~\ref{fig:synsamples}.

\begin{table*}[!ht]
	\centering
	\caption{Performance comparison of pre-trained models on various downstream tasks. We choose ViT-16/B for MAE due to the availability. [Keys: {\color{red}Best}, {\color{blue}Second best}]}
	\resizebox{0.9\linewidth}{!}{
		\begin{tabular}{|c|c|c|c|c|c|c|c|c|c|c|}
			\hline \multirow{4}{*}{Method} & \multirow{4}{*}{ Model - \# of params} & \multirow{4}{*}{Pre-train Datasets} & \multirow{4}{*}{Data Scale} & \multirow{4}{*}{Supervision} & \multicolumn{6}{|c|}{Downstream Tasks}   \\
			\cline{6-11}   & & &  & & \multicolumn{2}{|c|}{CelebA \cite{liu2015deep}} & \multicolumn{2}{|c|}{RAF-DB \cite{li2017reliable,li2019reliable}} & \multicolumn{2}{|c|}{ 300W \cite{sagonas2013300,sagonas2016300}}  \\
			\cline{6-11}  & & &  & & \multicolumn{2}{|c|}{mAcc. $\uparrow$}  & \multicolumn{2}{|c|}{mAcc. $\uparrow$} & \multicolumn{2}{|c|}{$\text{NME}_{\text{inter-ocular}}$ $\downarrow$} \\
			\cline{6-11}  & & & & & 100\% & 1\%  & 100\% & 10\% & 100\% & 10\% \\ 
			\hline DeiT  \cite{touvron2021training} & ViT-S/16 - 21M & ImageNet-1K  & 1.3M & images, labels & 90.79  & 88.27 &  87.87 & 75.36 & 3.40 & 4.34 \\	
			\hline MAE* \cite{he2022masked}  & ViT-B/16 - 86M & ImageNet-1K  & 1.3M & images & 91.16 &  \color{blue}{90.17} &  88.33 & 76.84 &  \color{blue}{3.36} &  \color{blue}{4.13} \\
			\hline DINO \cite{caron2021emerging} & ViT-S/16 - 21M & ImageNet-1K & 1.3M & images &  \color{blue}{91.25} & 89.62 & 88.23 & 75.85 & 3.53 & 4.57 \\
			\hline MSN \cite{assran2022masked} & ViT-S/16 - 21M & ImageNet-1K & 1.3M & images & 91.17 & 89.99 &  87.81 & 76.21  & 3.48 & 4.26  \\
			\hline FRL \cite{bulat2022pre} & ResNet50 - 23M & Large-Scale-Face  & 5M & face images & 91.04  & 90.04 &   \color{blue}{90.07} &  \color{blue}{80.57} & 3.85 & 4.25 \\
			% \hline DINO2 \cite{oquab2023dinov2} & ViT-S/14 - 21M & ImageNet-22K & 14M & images & 91.30 & 83.50 &   &  &  &   \\
			% \hline \ourmodel{}-1M-syn & MS1M & 1M & face images & \color{blue}{91.58} &  \color{blue}{90.47} &   89.06 &  80.11 &   \color{blue}{3.36}  &  4.24  \\
			% \hline \ourmodel{}-full-real & MS1M & 8.6M & face images & \color{red}{91.88} &  \color{red}{90.85} &   \color{red}{91.04} &  \color{red}{80.86} &   \color{red}{3.26}  &  \color{red}{4.04}   \\
			% \hline \hline \ourmodel{}-1M-syn &  ViT-S/16 - 21M & MS1M & 1M & face images & \color{blue}{91.57} &  \color{blue}{90.57} &  89.06  &  80.11 &   \color{blue}{3.36}  &  4.25  \\
			
			\hline \ourmodel{}-full-real & ViT-S/16 - 21M & MS1M & 8.6M & face images & \color{red}{91.88} &  \color{red}{90.86} &   \color{red}{91.04} &  \color{red}{82.10} &   \color{red}{3.27}  &  \color{red}{3.92}   \\	
			\hline
		\end{tabular}
	}
	\vskip -10 pt
	\label{tb:pretrain models}
\end{table*}

\subsection{Downstream tasks}

\noindent \textbf{Face attributes recognition} is a multi-class classification task that predicts multiple facial attributes ({\it e.g.}~gender, race, hair color) given one facial image. In this work, we evaluate the pre-trained model on two datasets: CelebA~\cite{liu2015deep} and LFWA \cite{huang2008labeled,liu2015deep}, containing 202,599 and 13,143 images respectively. 
They both have 40 annotated attributes per image. Following the protocols~\cite{liu2015deep,shu2021learning,zheng2022general} we use 162,770 images for training and 19,962 for testing on CelebA, and 6,263 images for training and the rest for testing on LFWA. Other than the initial weights from \ourmodel{}, we follow the same protocols and the average accuracy on all attributes is reported.

\noindent \textbf{Face expression recognition} is a single-class classification task that estimates one facial expression ({\it e.g.}  happy, anger, disguising) of a given face image. 
We evaluate the pre-trained model on two datasets:  RAF-DB \cite{li2019reliable,li2017reliable} and AffectNet \cite{mollahosseini2017affectnet}. RAF-DB contains around 29,670 face images from real-world databases, of which 15,339 images for 7 expression classifications.
There are 12,271 images for training and the remaining 3,068 for testing. 
AffectNet~\cite{mollahosseini2017affectnet} is a large-scale database for facial expressions. 
We use the most challenging AffectNet8 (including the additional "contempt" category) data, with 287,651 training images and 3,999 testing images. The average accuracy from all emotion classes is used as the evaluation metric.

\noindent \textbf{Face alignment} targets to regress the 2D coordinates of face landmarks on a face image. We evaluate our proposed model on two popular datasets: 300W \cite{sagonas2016300,sagonas2013300} and WFLW \cite{wu2018look}. 300W dataset contains 68 landmarks per face with 3,837 training images and 600 testing images. The WFLW dataset contains 68 landmarks as well,  with 7,500 training and 2,500 testing samples.  We measure the performance by the normalized mean error (NME).

\section{Experimental results}
%We compare the proposed \ourmodel{} with other general pre-training models and the state-of-the-art methods in each face analysis task respectively.
Besides the results in this section, more ablation studies, like "baseline methods pre-trained on face dataset" and "numbers of prototypes", can be found in \textit{supplementary documents}.

\subsection{Implementation}

After pre-training, the teacher model is used for downstream tasks training with additional head(s), both with end-to-end fine-tuning or head-only fine-tuning. For different tasks, the head designs are slightly varied. We donate the features from $h$-th head of $k$-th layer, including the last and intermediate layers, of the visual transformer as $feat_{k} = \{ f_{cls, k},  f_{1, k}, f_{2, k}, \cdots, f_{h, k}\}$, where $k=\{1,2, \cdots, 12\}$.

In particular, we use the multi-task classifiers \cite{cao2018partially} for face attributes classification. The cls-token feature vector from the last layer ($f_{cls, 12}$) is layer-normalized and appended with 40 separate linear layers to generate the logits for binary classification on each attribute. The model is trained with the averaged binary-cross-entropy loss on each head and is optimized by AdamW \cite{loshchilov2018decoupled}. We set the effective learning rate as 5e-4, weight decay as 0.05, and layer decay as 0.65. The learning rate decreases to zero in 100 epochs.

For face expression recognition, we use the original ViT-S/16 but change the last linear layer output dimension. Specifically, the output vector dimension for RAF-DB is set to 7, and AffectNet8 is set to 8. When fine-tuning, the learning rate is 5e-4, the weight decay is 0.5, and the layer decay is 0.65.  The total training epochs are 100 for RAF-DB and 10 for AffectNet8. Like previous settings \cite{wen2021distract}, we use the imbalanced data sampler for AffectNet8.

For face alignment, the ground-truth landmarks are rendered as Gaussian heat-map at a size of $128\times 128$ with $\sigma$ and values $\in [0,1]$ \cite{wang2019adaptive,huang2021adnet}. The non-cls tokens on layers $\{4, 6, 8, 12\}$ are utilized. To leverage the spatial distribution of these tokens, we reshape each to the 2D feature map of $14\times 14$. The UperNet \cite{xiao2018unified} is followed to fuse these feature maps from each layers to a final heat-map logits \cite{zheng2022general,liu2021swin}. Following the prior work~\cite{zheng2022general}, a simple soft-label cross-entropy loss is utilized for training the model. We use the AdamW with a learning rate of 0.01 and a weight decay of 1e-5.

\begin{table*}[!ht]
	\centering
	\caption{Comparison with baseline methods of  face attribute estimation on CelebA and LFWA datasets with limited data. The $\text{mAcc.}\uparrow$ is used as the evaluation metric. [Keys: \colorbox{lightgray}{SoTA}, {\color{red}Best}, {\color{blue}Second best} ]
	}
	\resizebox{0.82\textwidth}{!}{
		\begin{tabular}{|c|c|c|c|c|c|c|c|c|c|c|}
			\hline Dataset &  \multicolumn{5}{|c|}{CelebA \cite{liu2015deep}} & \multicolumn{5}{c|}{LFWA \cite{huang2008labeled,liu2015deep}} \\ 
			\hline Portion & 0.2\% & 0.5\% & 1\% & 2\% & 100\% & 5\% & 10\% & 20\% & 50\% & 100\%   \\
			\hline \# of training data & 325 & 843 & 1,627 & 3,255 & 162,770  & 313 & 626 & 1,252 &  3,131 & 6,263 \\
			\hline \rowcolor{gray!50}PS-MCNN \cite{cao2018partially} &- & -& -& -&  \color{red}{92.98} & -& -&- & -& \color{red}{87.36}   \\
			\hline Slim-CNN \cite{sharma2020slim} &79.90 & 80.20 & 80.96 & 82.32 & 91.24 & 70.90 & 71.49 & 72.12 & 73.45 & 76.02   \\
			\hline FixMath \cite{sohn2020fixmatch} & 80.22 & 84.19 & 85.77 & 86.14 & 89.78 & 71.42 & 72.78 & 75.10 & 80.87 & 83.84 \\
			\hline VAT \cite{miyato2018virtual} & 81.44 & 84.02 & 86.30 & 87.28 & 91.44 & 72.19 & 74.42 & 76.26 & 80.55 & 84.68 \\
			\hline SSPL\cite {shu2021learning} & 86.67  & 88.05 & 88.84  & 89.58 & 91.77& 78.68 & 81.65 & 83.45 & 85.43 & 86.53 \\
			\hline FARL \cite{zheng2022general} & 88.51 & 89.12 & $90.24$ & $90.55$  & \color{blue}{91.88} & 82.57 & $83.58$ & $84.80$ & $85.95$ & 86.69 \\
			\hline
			\hline \ourmodel{}-1M-syn & 88.60 & 89.78 & 90.57 & 90.92 & 91.57 & 82.69  & 83.92 &  \color{blue}{85.50} &  86.75 & 86.83 \\
			\hline \ourmodel{}-1M-real &  \color{blue}{88.70} & \color{blue}{90.15} & \color{blue}{90.72}  & \color{blue}{91.08} & 91.58 &  \color{blue}{82.73} &  \color{blue}{84.57} &  \color{black}{85.24} &  \color{blue}{86.79} &  \color{black}{87.06} \\
			\hline  \ourmodel{}-full-real & \color{red}{88.76} & \color{red}{90.43} & \color{red}{90.86}  & \color{red}{91.17}  & \color{blue}{91.88} & \color{red}{83.25} & \color{red}{85.13} & \color{red}{86.25}  & \color{red}{86.85}  & \color{blue}{87.08} \\   
			\hline
		\end{tabular}
	}
	\label{tb: face attributes}
\end{table*}

\subsection{Comparing with other pre-training models}

We clarify our models under different settings:
\begin{itemize}
	\item \ourmodel{}-1M-syn: pre-trained with 1M synthetic images from the generator, which is trained with randomly selected 1M real images.
	\item \ourmodel{}-1M-real: pre-trained with randomly selected 1M real images from MS1M dataset.
	\item \ourmodel{}-full-real: pre-trained with all real images from MS1M dataset.
\end{itemize}
We investigate how the pre-training models influence the downstream tasks' performance in Table~\ref{tb:pretrain models}. We compare the models from different architectures on various datasets. In particular, five different pre-training models are included: (1) DeiT \cite{touvron2021training}: was the improved ViT trained on ImageNet-1K with distillation under full supervision.  (2) MAE \cite{he2022masked}: was an auto-encoder learner for images reconstructed from masked input. It was trained on ImageNet-1K with self-supervision. (3) DINO \cite{caron2021emerging} was trained on ImageNet-1K as a form of mean teacher self-distillation under image self-supervision. (4) MSN \cite{assran2022masked} was a masked Siamese network trained on ImageNet-1K with self-supervised learning. (5) FRL~\footnote{https://github.com/1adrianb/unsupervised-face-representation}~\cite{bulat2022pre} trained ResNet50 \cite{he2016deep} on a large-scale face dataset without labels. For a fair comparison, we use ViT-S/16 as the backbone for DeiT, DINO, and MSN but ViT-B/16 for MAE due to the availability. We compare the proposed method with those baselines in the downstream tasks as illustrated in Table~\ref{tb:pretrain models}. As we can observe, all these models show a reasonable performance. The \ourmodel{}-full-real shows superior performance over both fully-supervised and self-supervised methods. The \ourmodel{}-1M-syn shows a competitive performance to the other baselines as well.

\subsection{Comparing with state-of-the-art face methods}
We compare our proposed model with other SoTA methods in both full-shot and few-shot settings in multiple downstream tasks.  All the input images are resized to $224\times 224$ and the official aligned version  (if applicable) is used. We conduct all the following experiments five times and report the average performance.

%\paragraph{Face attributes recognition}
\noindent \textbf{Face attributes recognition}
We compare our proposed method with baseline methods under both full-shot and few-shot settings. As we can observe in Table~\ref{tb: face attributes}, our proposed method shows superiority over all the other methods in few-shot and rank the $2$-nd under the full-shot with FaRL \cite{zheng2022general}. Note the PS-MCNN-LC \cite{cao2018partially} achieved a higher accuracy by using extra identity labels and a fine-grained network design to leverage the attribute relation. Meanwhile, we can observe that our method indeed benefits from a larger data scale (\ourmodel{}-1M-real {\it vs} \ourmodel{}-full-real). With only 1M synthetic data, \ourmodel{}-1M-syn also outperforms the baselines in all the few-shot settings. It is impressive to see that when trained with only 50\% data in LFWA, our proposed methods are still better than the close competitor FaRL \cite{zheng2022general} in 100\% data usage. When comparing results from \ourmodel{}-1M-syn {\it vs} \ourmodel{}-1M-real, the synthetic face data give competitive results with the one with real data in both settings. In general, the largest real data model \ourmodel{}-full-real outperforms both \ourmodel{}-1M-syn and \ourmodel{}-1M-real, which achieves the best results among all in the few-shot settings.
%Meanwhile, we can observe that \ourmodel{}-1M-syn outperforms the baselines in all the few-shot settings trained with only 1M synthetic data. In addition, we conduct another experiment for \ourmodel{}-1M-real. We can observe the model trained on the synthetic images achieves close performance to the one trained with real images. One thing surprising here is when trained with only 50\% data in LFWA, our proposed methods still perform better than the close competitor FaRL \cite{zheng2022general} in 100\% data usage.

\begin{table}[t]
	\centering
	\caption{Comparison with SoTA results of facial expression recognition on AffectNet8 and RAF-DB datasets. The $\text{mAcc.}\uparrow$ is used here as the evaluation metric. [Keys: \colorbox{lightgray}{SoTA}, {\color{red}Best}, {\color{blue}Second best}]}
	\resizebox{0.98\linewidth}{!}{
		\begin{tabular}{|c|c|c|c|c|c|c|c|}
			\hline  &  \multicolumn{3}{|c|}{AffectNet8 \cite{mollahosseini2017affectnet}} & \multicolumn{4}{|c|}{RAF-DB \cite{li2017reliable,li2019reliable}} \\
			\hline Methods & 2\% & 10\% & Full & 1\% & 2\% & 10\% & Full \\   
			\hline \rowcolor{gray!50} EAC \cite{zhang2022learn} & - & -  & 63.11 &   57.95 & 64.05 & \color{blue}{82.07}  & \color{blue}{89.99} \\
			\hline DAN \cite{wen2021distract} & 43.16 & \color{red}{52.41} & 62.09 & 53.17 & 58.46  & 78.05  & 89.70  \\
			\hline 
			% \hline \ourmodel{}-full-syn & \color{blue}{43.99} & 50.39 & \color{blue}{63.19}  & 89.43 & 80.30 & 66.46 & 59.11 \\
			\hline  \ourmodel{}-1m-syn & 43.46 & 49.96 & 62.59 & 58.74  & 66.13 & 80.11   &  89.06 \\
			\hline  \ourmodel{}-1M-real & 43.64 & 50.16 & \color{blue}{63.44}   & \color{blue}{61.04} & \color{blue}{67.60} & 80.32  & 89.83  \\
			\hline  \ourmodel{}-full-real & \color{red}{45.91} & \color{blue}{50.66} & \color{red}{63.64}  &  \color{red}{63.06} & \color{red}{70.61}  & \color{red}{82.10}  & \color{red}{91.04}  \\
			\hline
		\end{tabular}
	}
	\label{tb:face expression}
\end{table}

\begin{table*}[!htb]
	\centering
	\caption{Comparison of facial alignment on WFLW and 300W (test-set) dataset. The results of each column in 300W stand for the Common, Challenge, and Full subsets respectively.  The $\text{NME}_\text{inter-ocular} \downarrow$ is used here as the evaluation metric. [Keys: \colorbox{lightgray}{SoTA}, {\color{red}Best}, {\color{blue}Second best}]}
	\resizebox{0.82\linewidth}{!}{
		\begin{tabular}{|c|c|c|c|c|c|c|c|c|c|c|c|c|c|c|}
			\hline  & \multicolumn{5}{|c|}{WFLW \cite{wu2018look}} & \multicolumn{9}{|c|}{300W \cite{sagonas2013300,sagonas2016300}} \\
			\hline Methods & 0.7\% & 5\%& 10\%  &  20\% & 100\% & \multicolumn{3}{|c|}{1.5\%} & \multicolumn{3}{|c|}{10\%}  & \multicolumn{3}{|c|}{100\%}\\
			\hline \rowcolor{gray!50} FaRL \cite{zheng2022general} & 6.02 & 4.83 & 4.55 & 4.33 & 4.03 & \multicolumn{3}{|c|}{5.87 \ 3.24 \ 3.76} & \multicolumn{3}{|c|}{2.81 \ 4.83 \ 3.21}  & \multicolumn{3}{|c|}{2.56 \ 4.45 \ 2.93} \\
			\hline 
			\hline RCN+ \cite{honari2018improving} & -&- &- &-&- & \multicolumn{3}{|c|}{\threeemdash  \ \threeemdash  \ \threeemdash } & \multicolumn{3}{|c|}{\threeemdash  \  6.63 \ 4.47} & \multicolumn{3}{|c|}{3.00 \   4.98 \ 3.46} \\
			\hline SA \cite{qian2019aggregation} & - & - & 7.20 & 6.00 & \color{blue}{4.39} & \multicolumn{3}{|c|}{\threeemdash  \ \threeemdash  \ \threeemdash} & \multicolumn{3}{|c|}{ \threeemdash  \ \threeemdash  \ 4.27} & \multicolumn{3}{|c|}{3.21 \ 6.49 \ 3.86} \\
			\hline $\text{TS}^{3}$ \cite{dong2019teacher} &- & -&- &- & -& \multicolumn{3}{|c|}{ \threeemdash  \ \threeemdash \ \threeemdash} & \multicolumn{3}{|c|}{4.67 \ 9.26 \ 5.64} & \multicolumn{3}{|c|}{2.91 \ 5.90 \ 3.49} \\
			\hline 3FabRec \cite{browatzki20203fabrec} & 8.39 & 7.68 & 6.73 & 6.51 &  5.62 & \multicolumn{3}{|c|}{\color{blue}{4.55} \ \color{red}{7.39} \ \color{blue}{5.10}} & \multicolumn{3}{|c|}{3.88 \ 6.88 \ 4.47} & \multicolumn{3}{|c|}{3.36 \ 5.74 \ 3.82} \\
			\hline FSMA \cite{wei2021few} &- &- &- &- &- & \multicolumn{3}{|c|}{\threeemdash  \ \threeemdash  \ \threeemdash} & \multicolumn{3}{|c|}{3.59 \ 7.01 \ 4.45} & \multicolumn{3}{|c|}{3.12 \ 6.14 \  3.88} \\
			\hline FRL \cite{bulat2022pre} & \color{red}{7.11} & - & \color{blue}{5.44} & - & 4.57 & \multicolumn{3}{|c|}{\threeemdash  \ \threeemdash  \ \threeemdash}  & \multicolumn{3}{|c|}{\threeemdash  \ \threeemdash  \ 4.25} & \multicolumn{3}{|c|}{\threeemdash  \ \threeemdash  \ 3.85} \\
			\hline
			% \hline \ourmodel{}-1M-real & 8.21 & \color{blue}{6.08} & 5.56 & 5.15 & 4.53 & \multicolumn{3}{|c|}{ 4.58 \ 8.57 \ 5.36 } & \multicolumn{3}{|c|}{ \color{blue}{3.54} \ \color{blue}{6.24} \ \color{blue}{4.07} } & \multicolumn{3}{|c|}{ \color{blue}{2.93} \ \color{blue}{5.03} \ \color{blue}{3.35} } \\
			\hline \ourmodel{}-1M-syn  & 8.63 & 6.25 & 5.70 & 5.18 & 4.55 & \multicolumn{3}{|c|}{4.65 \ 8.89 \ 5.49} & \multicolumn{3}{|c|}{3.70 \ 6.57 \ 4.25 }& \multicolumn{3}{|c|}{2.95 \ 5.06 \ 3.36}   \\
			\hline \ourmodel{}-1M-real & 8.13 & \color{blue}{6.08} & 5.56 & 5.11 & 4.51 & \multicolumn{3}{|c|}{ 4.56 \ 8.55 \ 5.35 } & \multicolumn{3}{|c|}{ \color{blue}{3.54} \ \color{blue}{6.20} \ \color{blue}{4.06} } & \multicolumn{3}{|c|}{ \color{blue}{2.90} \ \color{blue}{5.05} \ \color{blue}{3.32} } \\
			\hline \ourmodel{}-full-real & \color{blue}{7.73} & \color{red}{5.75} & \color{red}{5.30} & \color{red}{4.90} & \color{red}{4.37} & \multicolumn{3}{|c|}{\color{red}{4.38} \ \color{blue}{7.88} \ \color{red}{5.08}} & \multicolumn{3}{|c|}{\color{red}{3.41} \ \color{red}{6.01} \ \color{red}{3.92}} & \multicolumn{3}{|c|}{\color{red}{2.86} \ \color{red}{4.92} \ \color{red}{3.27}} \\
			\hline
		\end{tabular}
	}
	\label{tb:alignment}
\end{table*}

%\paragraph{Face expression recognition} 
\noindent \textbf{Face expression recognition}
We conduct another set of experiments on facial expression recognition. Similar to previous experiments, we evaluate in both full-shot and few-shot settings. Due to the limited models for few-shot face expression recognition, we compare the proposed models with two SoTA baselines: DAN \cite{wen2021distract} (ResNet-18) and EAC \cite{zhang2022learn} (ResNet-50). For the full-shot results, the results are copied from the paper report.   For the few-shot results, we run the experiments with their published codes \footnote{https://github.com/yaoing/DAN}\footnote{https://github.com/zyh-uaiaaaa/Erasing-Attention-Consistency}. For a fair comparison, both DAN and EAC models were initialized with trained weights from fully-supervised training on MS1M dataset.  As shown in Table~\ref{tb:face expression}, the \ourmodel{}-full-real model achieves a better performance than the EAC and DAN methods in both RAF-DB and AffectNet8 datasets under both full and limited data settings.  Similarly, we can observe the larger pre-training data gives better performance (\ourmodel{}-full-real {\it vs} \ourmodel{}-1M-real). When comparing the AffectNet8 results of DAN using 100\% and 10\% training data to the ones from our \ourmodel{}, one reason for the degradation on 100\% could be the long-tail bias \cite{liu2016large,ouyang2016factors} from the MS1M weights in fully-supervised training.

%\paragraph{Face alignment}
\noindent \textbf{Face alignment}
We also evaluate the \ourmodel{} models on face alignment tasks using WFLW and 300W test-set. As shown in Table~\ref{tb:alignment}, our \ourmodel{}-full-real model surpasses all baselines under the full settings and most under the few-shot settings.  We also include the SoTA (FaRL \cite{zheng2022general}) method in this table, which was trained in semi-supervision on 20M  web-text and image pairs while we are self-supervision on fewer face images only. For the close competitor FRL~\cite{bulat2022pre}, it only surpasses ours under the  0.7\%  few-shot setting on WFLW. When comparing \ourmodel{}-1M-real and \ourmodel{}-1M-syn, we can see using the real face images leads to a better result.

\begin{figure*}[t]
	\centering
	\begin{subfigure}[b]{0.4\linewidth}
		\centering
		\includegraphics[width=\linewidth]{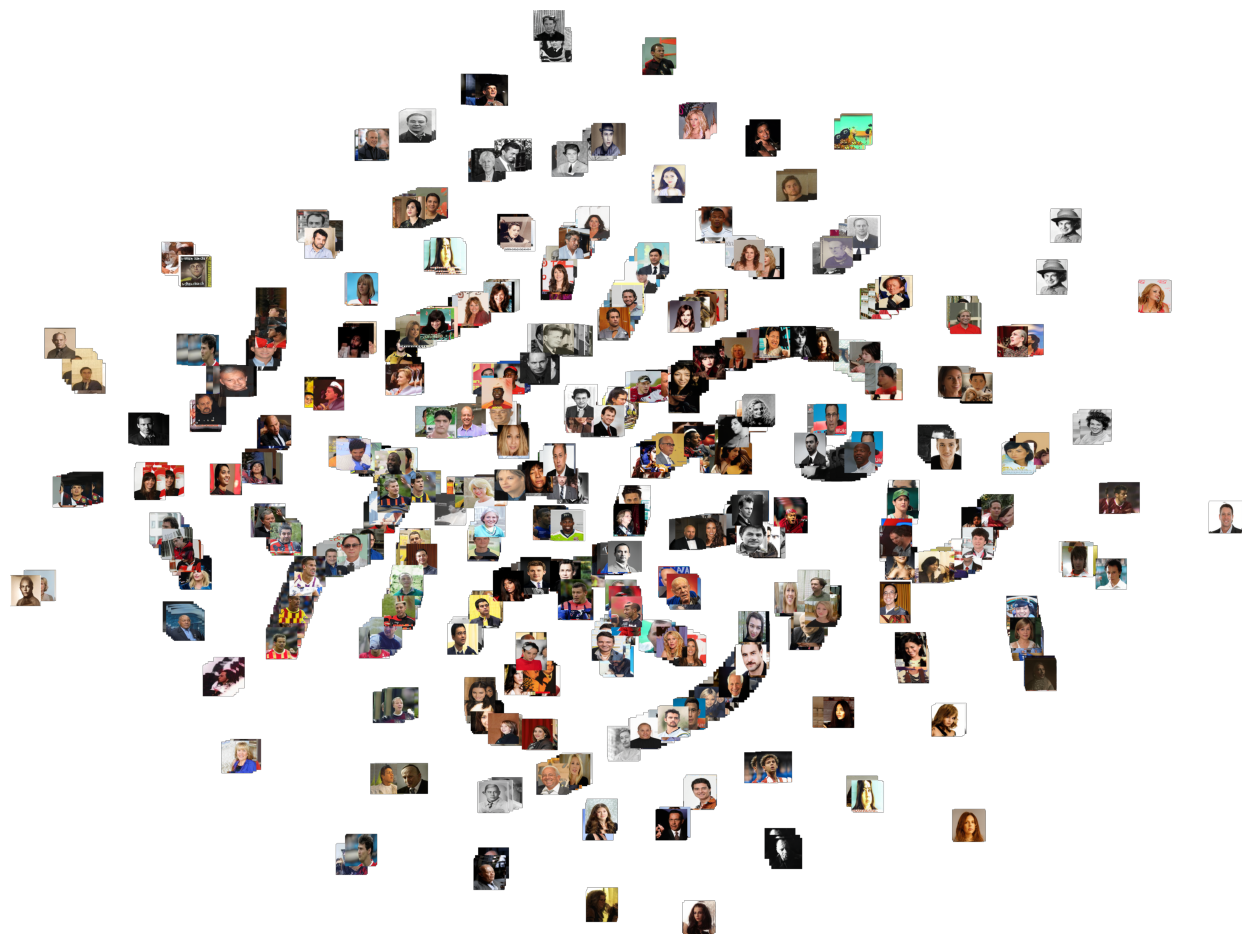}
		\caption{1M real images}\label{fig:tsne_1M}
	\end{subfigure}
	\begin{subfigure}[b]{0.4\linewidth}
		\centering
		\includegraphics[width=\linewidth]{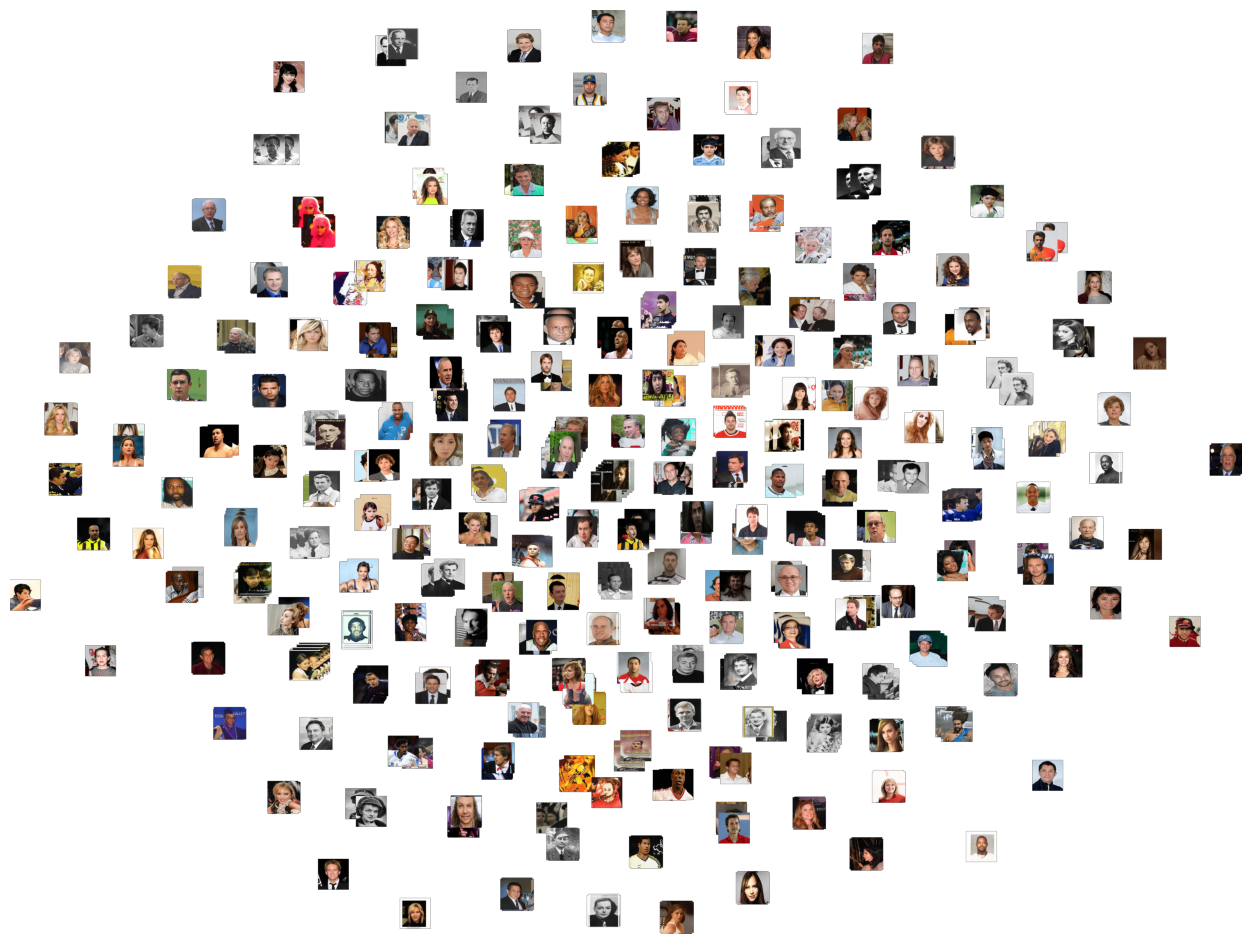}
		\caption{Full real images}\label{fig:tsne_8M}
	\end{subfigure}
	\caption{t-SNE visualization \cite{van2008visualizing} of learned prototypes by finding the nearest neighbor in synthesized training face images.}
	\label{fig:tsneimages}
\end{figure*}

\subsection{Visualization of learned prototypes}

% To understand what variances are covered by the learned prototypes, we conduct a t-SNE visualization \cite{van2008visualizing} on the $1024$ prototypes of \ourmodel{}-1M-real and \ourmodel{}-full-real. In particular, we conduct the nearest neighbor retrieval between the prototypes and training images. As we can observe in Figure~\ref{fig:tsneimages}, the learned prototypes are mostly scattered in a sparse distribution, which matches the self-supervised learning idea that each image is treated as an individual class. 
In order to exhibit the variances captured by the learned prototypes, we perform a t-SNE visualization \cite{van2008visualizing} on the set of 1024 prototypes derived from the ProS-1M-real and ProS-full-real. Specifically, we employ nearest-neighbor retrieval to establish connections between the learned prototypes and the training images. Figure~\ref{fig:tsneimages} illustrates the results, indicating that the learned prototypes are predominantly dispersed in a sparse distribution. Comparing with Figure~\ref{fig:tsne_1M} and Figure~\ref{fig:tsne_8M}, we can observe that the learned prototypes show a better span coverage when using the full size of the training dataset.

\begin{figure}[!ht]
	\centering
	\begin{subfigure}[!]{0.98\linewidth}
		\centering
		\includegraphics[width=0.24\linewidth]{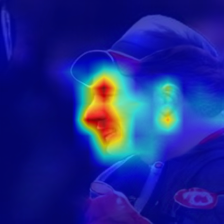}\hfill
		\includegraphics[width=0.24\linewidth]{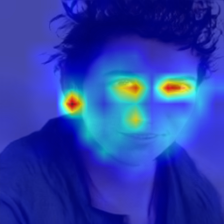}\hfill
		\includegraphics[width=0.24\linewidth]{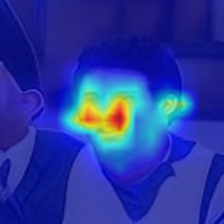}\hfill
		\includegraphics[width=0.24\linewidth]{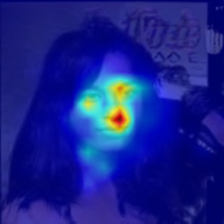}\hfill
		\caption{Pre-trained on MS1M \cite{Guo2016MSCeleb1MAD}}
	\end{subfigure}
	\hfill
	\begin{subfigure}[!]{0.49\linewidth}
		\centering
		\includegraphics[width=0.5\linewidth]{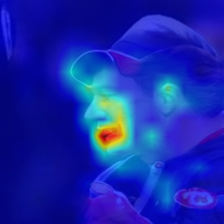}\hfill
		\includegraphics[width=0.5\linewidth]{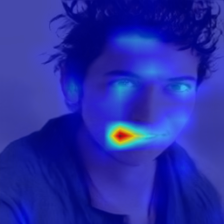}\hfill
		\caption{Fine-tuned on CelebA \cite{liu2015deep}}
	\end{subfigure}
	\begin{subfigure}[!]{0.49\linewidth}
		\centering
		\includegraphics[width=0.5\linewidth]{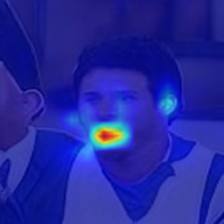}\hfill
		\includegraphics[width=0.5\linewidth]{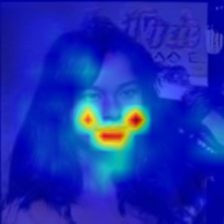}\hfill
		\caption{Fine-tuned on RAF-DB \cite{li2017reliable,li2019reliable}}
	\end{subfigure}
	\caption{Comparison of attention heatmaps of teacher models from pre-trained (a) on MS1M \cite{Guo2016MSCeleb1MAD}; and fine-tuned (b) on CelebA \cite{liu2015deep} and (c) on RAF-DB \cite{li2017reliable,li2019reliable} respectively. }
	\label{fig:attention heatmap}
\end{figure}

%\paragraph{Visualization of pretrained model}
\subsection{Visualization of pre-training model}
% We study what the proposed \ourmodel{} model learned from the pre-training and fine-tuning by visualizing the attention mechanism in heatmaps as shown in Figure~\ref{fig:attention heatmap}. We use the teacher model of \ourmodel{}-full-real and the heat-map value is summed up from attention heads of the last self-attention layer. For pre-training,  our \ourmodel{} model can successfully localize the facial region of the input images with different variations (pose, scale, {\it etc.}) showing high similarity with human attention. By comparing heatmaps from the pre-trained model and fine-tuned models, we find that the features learned from fine-tuned models are more specific to the downstream tasks, while the features learned from the pre-trained model are more general.
In order to investigate the knowledge acquired through the pre-training and fine-tuning stages of our proposed \ourmodel{} model, we visualize the attention mechanism using heatmaps, as depicted in Figure~\ref{fig:attention heatmap}. Specifically, we use the teacher model of \ourmodel{}-full-real and analyze the heatmap values obtained from the attention heads in the final self-attention layer.  During the pre-training phase, our \ourmodel{} model successfully localizes the facial region in input images, effectively capturing various variations such as pose and scale. These heatmaps exhibit a similar pattern to human attention patterns. Comparing the heatmaps generated by the pre-trained model and the fine-tuned models, we observe that the features learned through fine-tuning become more task-specific, while the features derived from the pre-trained model maintain a more generalized representation.

\section{Discussion}
\noindent \textbf{Ethical considerations}
For training, \ourmodel{} requires large-scale face images. We use the MS1M data, one of the largest datasets in the world, collected by Microsoft in 2016. The images in this dataset are scraped off the web under the terms of the Creative Commons license and are limited to the academic usage of the photos. 
Studies \cite{zhu2022webface260m} show that the MS1M dataset is subject to biases due to label noise, duplicate images, and non-face images. It is essential to acknowledge and address these biases to prevent the propagation of unfair or discriminatory practices in the development and deployment of face recognition systems.

\noindent \textbf{Social Impact and Limitation} We have identified certain limitations in the pre-training phase of our \ourmodel{} model. Specifically, we found that the bias issue present in the MS1M dataset is still present in the learned representations of \ourmodel{}. Moreover, our model does not perform exceptionally well in face parsing tasks on the LaPa dataset \cite{liu2020new}, as evidenced in the supplementary documents. One possible explanation for this performance gap is that the learned features in \ourmodel{} tend to be more semantic-specific rather than spatial-specific. This observation is further supported by Figure~\ref{fig:attention heatmap}, where we can observe that the attention mechanism does not adequately attend to the hair region.

% We believe during the pre-training, the bias issue in MS1M dataset is still inherited in the \ourmodel{}. In addition, \ourmodel{} fails to achieve excellent results on the face parsing on LaPa dataset \cite{liu2020new} as shown in supplementary documents. It could be the reason that the learned features are semantic-specific instead of spatial-specific. This could also be observed in Figure~\ref{fig:attention heatmap} that the attention mechanism does not pay interest to the hair part. 
% One reason could be that the learned features mostly cover the facial region but not the hair region as indicated in Figure~\ref{fig:attention heatmap}.

\section{Conclusion}
In this paper, we present a self-supervised pre-training method (\ourmodel{}) for learning face representation from unlabeled large-scale images only. One modified prototype-based matching loss and a face-aware retrieval system are introduced along with the \ourmodel{}. We explore the \ourmodel{} on both real and synthetic face images. In addition, we show the face representations learned from \ourmodel{} can be well-transferred to multiple downstream face analysis tasks including attribute estimation, expression recognition, and face alignment. Compared with SoTA methods, our proposed \ourmodel{}  shows the superiority of performance in the limited data. Moreover, the proposed method surpasses the previous SoTA methods in facial expression recognition.

%%-------------------------------------------------------------------------

\clearpage

%%%%%%%%% REFERENCES
{\small
	\bibliographystyle{ieee_fullname} 
	\bibliography{face}
}

\clearpage
\appendix

\appendix

\paragraph{Appendices} 
In this supplementary material, we conduct a series of studies on the proposed models as follows.

\section{More ablation study}

\noindent \textbf{Study on face-aware retrieval system}
We evaluate the model performances with/without the face-aware retrieval system as shown in Table~\ref{tb: retrieval study}. As we can see, the face-aware retrieval improves the model performances on all three tasks.

\begin{table}[!ht]
	\centering
	\caption{Study on face-aware retrieval system.}
	\resizebox{0.9\linewidth}{!}{
		\begin{tabular}{|c|c|c|c|c|}
			\hline
			Model & & CelebA & RAF-DB & 300W \\
			\hline
			\multirow{2}{*}{\ourmodel{}-1M-real} & w/o & 91.42 & 89.34 &  3.35 \\
			& w/ (ours) & 91.58 & 89.83 &  3.32 \\
			\hline
			%   \multirow{2}{*}{\ourmodel{}-full-real} & w/o & 91.88 & 91.04 &  3.27 \\
			% & w/ (ours) & 91.88 & 91.04 &  3.26 \\
			% \hline
		\end{tabular}
	}
	\label{tb: retrieval study}
\end{table}
\noindent \textbf{The different number of prototypes, architecture and training time:} 
We compare the performances of the proposed \ourmodel{}-1M-syn model on the different numbers of prototypes, architecture, and training epochs.
The results are shown in Table~\ref{tb:variant training}.  As we can observe, the performances are improved with the 
% increasing number of prototypes from 1,512,1024 and start 
increasing number of prototypes from 1 \footnote{we use the loss in Dino\cite{caron2021emerging}}, 512,1024 and start degrading at 2048. Therefore, we set the default number of prototypes as 1024. In addition, we evaluate the model with a longer training time (100 {\it vs} 20 epochs) and a larger model ViT-B/16 (85M {\it vs} 21M). We can observe the longer training iterations and a larger model size do slightly improve the model performances.

\begin{table}[!ht]
	\centering
	\caption{Ablation study of different number of prototypes, training epochs and model architecture on \ourmodel{}-1M-syn, which is trained on 1024 prototypes, 20 epochs and ViT-S/16.}
	\resizebox{0.9\linewidth}{!}{
		\begin{tabular}{|c|c|c|c|c|}
			\hline
			& & CelebA & RAF-DB & 300W \\
			\hline
			\multirow{4}{*}{\# of prototypes} 
			& 1 & 90.45 & 86.48 & 3.71 \\
			& 512 & 91.46 & 88.46 & 3.38 \\
			& 1024 (ours) & 91.57 & 89.06 &  3.36 \\
			% 1,024 (ours) & \textbf{91.57} & \textbf{89.06} &  \textbf{3.36} \\
			& 2,048 & 91.53 & 88.85 & 3.38  \\            
			\hline
			\multirow{2}{*}{epochs}
			& 20 (ours) & 91.57 & 89.06 &  3.36 \\
			& 100  & 91.59 & 89.44 &  3.35 \\
			\hline
			\multirow{2}{*}{architectures}
			& ViT-S/16 (ours) & 91.57 & 89.06 &  3.36 \\
			& ViT-B/16 & 91.52  & 89.53 &  3.35 \\ 
			\hline
		\end{tabular}
	}
	\label{tb:variant training}
\end{table}

\noindent \textbf{Data size:}
We study how the data size of face images could influence the final performance. In particular, we study the training data size of 0.2M, 0.5M, 1M, and 8M on real images. We report the results in Table~\ref{tb:data size}. As we can observe, the more training images we use, the better performance. 

\begin{table}[!ht]
	\centering
	\caption{Study on data size.}
	\resizebox{0.8\linewidth}{!}{
		\begin{tabular}{|c|c|c|c|}
			\hline Size & CelebA & 300W & RAF-DB \\
			\hline  0.2M & 91.45 & 3.57 & 81.75 \\
			\hline 0.5M & 91.53 & 3.48 & 85.91  \\
			\hline 1M  & 91.58 & 3.32 & 89.83  \\
			\hline 8.6M (full)  & 91.88 & 3.27 & 91.04  \\ 
			\hline
		\end{tabular}
	}
	\label{tb:data size}
\end{table}

\section{Models comparison}
The differences between the proposed method and existing ones \cite{caron2021emerging,caron2020unsupervised} are shown in Table~\ref{tb:comparison with swav and dino}. Compared with DINO, we add the prototypes and use the Sinkhorn regularization \cite{cuturi2013sinkhorn}. Compared with SwAV, we explore the momentum encoder and vision transformer architecture.

\begin{table}[!ht]
	\centering
	\caption{Comparison between proposed \ourmodel, DINO \cite{caron2021emerging} and SwAV \cite{caron2020unsupervised}}
	\resizebox{0.98\linewidth}{!}{
		\begin{tabular}{|c|c|c|c|c|c|}
			\hline
			Methods &  Momentum & Prototype & Operation (teacher) & Architecture & Dataset  \\
			\hline
			SwAV \cite{caron2020unsupervised} & & \checkmark & Sinkhorn \cite{cuturi2013sinkhorn} & ResNet & ImageNet\\
			\hline
			DINO \cite{caron2021emerging} & \checkmark & & Centering & Vision Transformer & ImageNet \\
			\hline 
			\ourmodel (ours) & \checkmark & \checkmark & Sinkhorn \cite{cuturi2013sinkhorn} & Vision Transformer & MS1M \\
			\hline
		\end{tabular}
	}
	\label{tb:comparison with swav and dino}
\end{table}

\subsection{Pre-training models on face dataset}

We re-implement the pre-training methods such as DINO \cite{caron2021emerging}, MAE \cite{he2022masked}, and MSN \cite{assran2022masked} models on the synthetic 1M images and evaluate the downstream tasks as shown in Table~\ref{tb:pretrain on face study}. For a fair comparison, we use the ViT-S/16 architecture for these methods and linearly scale the learning rate based on the data size. As we can observe, \ourmodel{} still outperforms the other baselines, especially on the expression estimation task at RAF-DB dataset. This indicates the superiority of the proposed method compared with the other baselines when trained with the same face dataset.

\begin{table}[h]
	\centering
	\caption{Experimental comparison with DINO \cite{caron2021emerging}, MAE \cite{he2022masked}, and MSN \cite{assran2022masked} methods on facial datasets }
	\resizebox{0.98\linewidth}{!}{
		\begin{tabular}{|c|c|c|c|}
			\hline
			Methods &  CelebA & RAF-DB & 300W \\
			\hline
			DINO \cite{caron2021emerging} & 91.45 & 87.48 & 3.41 \\
			MAE \cite{he2022masked} & 91.28 & 87.73 & 3.38 \\
			MSN \cite{assran2022masked} & 91.43 & 88.19 & 3.38 \\
			\ourmodel-1M-syn (ours) & 91.57 & 89.06 &  3.36 \\
			\hline
		\end{tabular}
	}
	\label{tb:pretrain on face study}
\end{table}

\section{Linear probe}\label{appendix:lp}
We analyze the feature learned from \ourmodel{}-1M-syn model by fine-tuning with frozen vision-transformer backbone and 
the study results are shown in Table~\ref{tb:lp}. As we can observe, the linear probe results from synthetic data are better on face attribute estimation. While, the model from real images achieves better performance on expression classification and face alignment.

\begin{table}[t]
	\centering
	\caption{Study on linear probe with frozen ViT-S/16 backbone.}
	\begin{subtable}{1.0\linewidth}
		\resizebox{0.98\textwidth}{!}{
			\begin{tabular}{|c|c|c|c|c|c|c|c|c|c|c|}
				\hline Dataset &  \multicolumn{5}{|c|}{CelebA} & \multicolumn{5}{c|}{LFWA} \\ 
				\hline Portion & 0.2\% & 0.5\% & 1\% & 2\% & 100\% & 5\% & 10\% & 20\% & 50\% & 100\%   \\
				\hline \# of training data & 325 & 843 & 1,627 & 3,255 & 162,770  & 313 & 626 & 1,252 &  3,131 & 6,263 \\
				% 			\hline \# of training data & 325 & 843 & 1,627 & 3,255 & 162,770  & 313 & 626 & 1,252 &  3,131 & 6,263 \\
				%				\hline \hline $\text{\ourmodel{}-1M-syn}_{lp}$ &  88.21 & 89.54 & 89.85 & 89.93 & 90.23& 82.55 & 83.26 & 83.98 & 84.76 & 85.14 \\
				\hline \hline $\text{\ourmodel{}-1M-syn}_{lp}$ &  87.42 & 88.64 & 89.17 & 89.67 & 90.23& 82.55 & 83.26 & 83.98 & 84.76 & 85.14 \\
				% \hline \ourmodel{}-1M-syn & 88.59 & 89.76 &90.47 & 90.92 & 91.58 & 82.77  & 83.91 &  85.50 &  86.75 & 86.80 \\
				%				\hline  \ourmodel{}-full-real & 88.75 & 90.40 & 90.85  & 91.15  & 91.88 & 83.22 & 85.03 & 86.24  & 86.86 & 86.89 \\
				\hline \hline $\text{\ourmodel{}-1M-real}_{lp}$ & 87.30 & 88.24 & 88.80 & 89.31 & 90.62 & 81.02 & 82.13 & 83.02 & 84.08 & 84.72 \\
				% \hline \ourmodel{}-1M-real & 88.43  & 89.77 & 90.45 & 90.90 & 91.52 &  82.73 &  84.61 &  85.87 &  86.80 &  87.07 \\
				\hline
			\end{tabular}
		}
	\end{subtable}
	\hfill
	\begin{subtable}{1.0\linewidth}
		\resizebox{0.98\textwidth}{!}{
			\begin{tabular}{|c|c|c|c|c|c|c|c|}
				\hline  &  \multicolumn{3}{|c|}{AffectNet8 } & \multicolumn{4}{|c|}{RAF-DB } \\
				\hline Methods & Full & 10\% & 2\% & Full & 10\% & 2\% & 1\% \\
				\hline \hline  $\text{\ourmodel{}-1m-syn}_{lp}$ & 42.06 & 38.48 & 33.78 & 80.04 & 73.40 &  64.86 & 56.23  \\
				% \hline  \ourmodel{}-1m-syn & 62.59 & 49.96 & 43.46 & 89.06 & 80.11 &  66.13 & 58.74 \\
				\hline \hline  $\text{\ourmodel{}-1m-real}_{lp}$ & 43.01 & 40.56 & 37.56  &  75.46 & 69.20 &  60.07 & 55.64  \\
				% \hline  \ourmodel{}-1M-real & 63.14 & 50.16 & 43.64  & 89.34 & 80.32 & 65.74 & 59.42 \\
				\hline
			\end{tabular}
		}
	\end{subtable}
	\hfill
	\begin{subtable}{1.0\linewidth}
		\resizebox{0.98\textwidth}{!}{
			\begin{tabular}{|c|c|c|c|c|c|c|c|c|c|c|c|c|c|c|}
				\hline  & \multicolumn{5}{|c|}{WFLW} & \multicolumn{9}{|c|}{300W} \\
				\hline \hline Methods & 0.7\% & 5\%& 10\%  &  20\% & 100\% & \multicolumn{3}{|c|}{1.5\%} & \multicolumn{3}{|c|}{10\%}  & \multicolumn{3}{|c|}{100\%}\\
				\hline $\text{\ourmodel{}-1M-syn}_{lp}$ &  10.73 &  8.00 & 7.39 & 6.94 & 6.12 & \multicolumn{3}{|c|}{ 5.56 \  11.12 \ 6.64 } & \multicolumn{3}{|c|}{ 4.32 \ 8.33 \ 5.12 } & \multicolumn{3}{|c|}{ 3.66 \ 6.72 \ 4.26 } \\
				% \hline \ourmodel{}-1M-syn  & 8.69 & 6.25 & 5.70 & 5.17 & 4.55 & \multicolumn{3}{|c|}{4.68 \ 8.93 \ 5.51} & \multicolumn{3}{|c|}{3.69 \ 6.56 \ 4.24} & \multicolumn{3}{|c|}{2.95 \ 5.05 \ 3.36}   \\
				\hline \hline $\text{\ourmodel{}-1M-real}_{lp}$ & 9.47 & 7.25 & 6.76 & 6.35 & 5.68 & \multicolumn{3}{|c|}{ 5.31 \ 10.44 \ 6.32 } & \multicolumn{3}{|c|}{ 4.17 \ 7.90 \ 4.90 } & \multicolumn{3}{|c|}{ 3.58 \ 6.39 \ 4.13 } \\
				% \hline \ourmodel{}-1M-real & 8.21 & 6.08 & 5.56 & 5.15 & 4.53 & \multicolumn{3}{|c|}{ 4.58 \ 8.57 \ 5.36 } & \multicolumn{3}{|c|}{ 3.54 \ 6.24 \ 4.07 } & \multicolumn{3}{|c|}{ 2.93 \ 5.03 \ 3.35 } \\
				\hline
			\end{tabular}
		}
	\end{subtable}	
	\label{tb:lp}
\end{table}

\noindent \textbf{Experiments on face parsing}
As shown in Table~\ref{tb:faceparsing}, \ourmodel{} fails to achieve excellent results on the face parsing on LaPa dataset. One reason could be that the learned features mostly cover the facial region but not the hair region, which can also be observed in the parsing result in the "Hair" class.
\begin{table}[t]
	\centering
	\caption{Comparison with SOTA  methods on LaPa \cite{liu2020new} dataset.}
	\resizebox{0.98\linewidth}{!}{
		\begin{tabular}{|c|c|c|c|c|c|c|c|c|c|c|c|}
			\hline  Subset & Skin & Hair & L-E & R-E & U-L & I-M & L-L & Nose & L-B & R-B & Mean  % & 10\% 
			\\
			\hline FaRL \cite{zheng2022general} & 97.52 & 95.11 & 92.33 & 92.09 & 88.69 & 90.70 & 90.05 & 97.55 & 91.57 & 91.34 & 92.70  % & 92.40 
			\\
			\hline AGRNet \cite{te2021agrnet} & 97.7 & 96.5 & 91.6 & 91.1 & 88.5 & 90.7 & 90.1 & 97.3 & 89.9 & 90.0 & 92.3  %  & 74.46 
			\\
			\hline 
			%			\hline  \ourmodel{}-1M-syn  & 96.95 & 93.20 & 91.09 & 90.86 & 87.58 & 89.47 & 89.26 & 97.45 & 90.47 & 89.60 & 91.60  &  88.55 \\
			%			\hline  \ourmodel{}-1M-real  & 97.05 & 93.55 & 91.02 & 91.20 & 88.01 & 89.73 & 89.26 & 97.40 & 90.34 & 89.95 & 91.70 % & 89.06  
			%			\\
			\hline  \ourmodel{}-1M-syn  & 96.95 & 93.20 & 91.09 & 90.86 & 87.58 & 89.47 & 89.26 & 97.45 & 90.47 & 89.60 & 91.60 %& 88.55 
			\\
			% 			\hline  \ourmodel{}-full-syn  & 97.10 & 93.70 & 91.31 & 90.86 & 88.11 & 90.00 & 89.62 & 97.49 & 90.77 & 89.99 & 91.90 & 88.90 \\
			\hline  \ourmodel{}-1M-real  & 97.05 & 93.55 & 91.02 & 91.20 & 88.01 & 89.73 & 89.26 & 97.40 & 90.34 & 89.95 & 91.70 % & 89.06  
			\\
			\hline  \ourmodel{}-full-real  & 97.13 & 93.57 & 91.42 & 91.32 & 88.27 & 90.10 & 89.51 & 97.52 & 90.88 & 90.27 & 92.00 % & 89.45 
			\\
			\hline
		\end{tabular}
	}
	\label{tb:faceparsing}
\end{table}

\end{document}